\documentclass{article}
\usepackage[margin=1.5in]{geometry}
\usepackage{amsmath}
\usepackage{amsthm}
\usepackage{algorithm, algorithmic}

\usepackage{amssymb}
\usepackage{url}
\usepackage{enumerate}
\usepackage{sidecap}

\usepackage{stmaryrd}
\usepackage{textcomp}

\usepackage[utf8]{inputenc} 
\usepackage[T1]{fontenc}    
\usepackage{hyperref}       
\usepackage{amsfonts}       
\usepackage{nicefrac}       
\usepackage{microtype}      

\usepackage[usenames, dvipsnames]{color}
\usepackage{stmaryrd}
\usepackage{textcomp}

\usepackage{graphicx}

\newcommand{\nc}[1]{\newcommand{#1}}
\nc{\bs}[1]{\boldsymbol{#1}}
\nc{\be}{\begin{equation*}}
\nc{\ee}{\end{equation*}}
\nc{\indi}[1]{\llbracket#1\rrbracket}
\nc{\emp}[1]{\emph{#1}}
\nc{\ips}{\mathcal{I}}
\nc{\tks}{\mathcal{X}}
\nc{\sps}{\mathcal{Z}}
\nc{\ops}{\mathcal{Y}}
\nc{\lss}{\mathcal{L}}
\nc{\art}{x}
\nc{\pd}{p}
\nc{\real}{\mathbb{R}}
\nc{\arl}{\ell}
\nc{\ari}{\omega}
\nc{\opf}{y}
\nc{\expt}[1]{\mathbb{E}[#1]}
\nc{\lost}[1]{\ell_{#1}}
\nc{\inst}[1]{\omega_{#1}}
\nc{\bt}{\mathcal{B}}
\nc{\rot}{r}
\nc{\des}[1]{{\Downarrow}(#1)}
\nc{\ver}{\mathcal{V}}
\nc{\vrs}{\mathcal{U}}
\nc{\res}[2]{\langle#1|\,#2\rangle}
\nc{\pss}{\mathcal{S}}
\nc{\css}{\mathcal{S}^\dag}
\nc{\pff}{f}
\nc{\cff}{f^\dag}
\nc{\pmf}{m}
\nc{\cmf}{m'}
\nc{\cvs}{\mathcal{C}}
\nc{\pnt}[1]{#1^{\uparrow}}
\nc{\lch}[1]{#1^{\triangleleft}}
\nc{\rch}[1]{#1^{\triangleright}}
\nc{\sib}[1]{#1^{\flat}}
\nc{\nec}{\epsilon}
\nc{\blank}{\circ}
\nc{\ocf}{q}
\nc{\gfun}{g}
\nc{\arm}{a}
\nc{\armp}{b}
\nc{\armpp}{c}
\nc{\yhat}[1]{\hat{y}_{#1}}
\nc{\prs}{\mathcal{P}}
\nc{\ppr}{\phi}
\nc{\cpr}{\phi^\dag}
\nc{\trr}{\psi}
\nc{\crn}{\gamma}
\nc{\ext}{\xi}
\nc{\as}{s}
\nc{\asp}{s'}
\nc{\ash}{\hat{s}}
\nc{\asph}{\hat{\sigma}}
\nc{\apr}{\theta}
\nc{\pat}[2]{\theta_{#1}(#2)}
\nc{\sprs}{\mathcal{Q}}
\nc{\crw}{\gamma}
\nc{\thh}[1]{\lambda_{#1}}
\nc{\pms}[2]{\mu_{#1}(#2)}
\nc{\cms}[2]{\mu'_{#1}(#2)}
\nc{\ai}{\iota}
\nc{\der}[1]{\nabla_{#1}}
\nc{\lr}{\eta}
\nc{\aro}{y'}
\nc{\cexd}[3]{\mathbb{E}_{#1}[#2\,|\,#3]}
\nc{\draw}{\sim}
\nc{\alg}{\textsc{Xprop}}
\nc{\la}{\leftarrow}
\nc{\lot}[2]{z_{#1}(#2)}
\nc{\arbt}{\tau}
\nc{\cvy}{\tilde{y}}
\nc{\pfft}[1]{\pff_{#1}}
\nc{\cfft}[1]{\cff_{#1}}
\nc{\pmft}[1]{\pmf_{#1}}
\nc{\cmft}[1]{\cmf_{#1}}
\nc{\gf}{g}
\nc{\gft}[1]{\gf_{#1}}
\nc{\hf}{\beta}
\nc{\arih}{\hat{\ari}}
\nc{\ecs}{\mathcal{E}}
\nc{\ecl}{\mathcal{C}}
\nc{\arn}{q}
\nc{\lof}{y^\ast}
\nc{\eqr}{\equiv}
\nc{\loft}[1]{y^\ast_{#1}}
\nc{\cmp}[1]{h_{#1}}
\nc{\ccm}{h}
\nc{\tbe}{\tilde{\hf}}
\nc{\imi}{\delta}
\nc{\imip}{\delta'}
\nc{\sln}{N}
\nc{\nat}{\mathbb{N}}
\nc{\ipn}[1]{\iota_{#1}}
\nc{\bse}{\bs{z}}
\nc{\se}[1]{z_{#1}}
\nc{\bset}{\bs{z}}
\nc{\set}[2]{z_{#1,#2}}
\nc{\ix}{i}
\nc{\jx}{j}
\nc{\kx}{k}
\nc{\ovl}{\zeta}
\nc{\vea}{\mathcal{W}}
\nc{\veb}{\mathcal{W}'}
\nc{\hr}[1]{a(#1)}
\nc{\hrp}[1]{a'(#1)}
\nc{\vr}[1]{b(#1)}
\nc{\vrp}[1]{b'(#1)}
\nc{\spa}{~,~}
\nc{\arc}{\mathcal{A}}
\nc{\seq}[2]{\langle#1\,|\,#2\rangle}
\nc{\arnn}{n}
\nc{\ins}{\mathcal{X}}
\nc{\nint}[1]{x_{#1}}
\nc{\nopt}[1]{y_{#1}}
\nc{\cexpt}[2]{\mathbb{E}[#1\,|\,#2]}
\nc{\pp}{\mathcal{P}}
\nc{\cp}{\mathcal{P}^\dag}
\nc{\tp}{\mathcal{Q}}
\nc{\tokf}{\tau}
\nc{\pars}[1]{{\Uparrow}(#1)}
\nc{\aps}{\mu}
\nc{\acs}{\hat{\mu}}
\nc{\app}{\theta}
\nc{\ppi}[2]{\theta_{#1}(#2)}
\nc{\tpi}[2]{\lambda_{#1}(#2)}
\nc{\cpi}[3]{\theta^\dag_{#1}(#2,#3)}
\nc{\pme}[2]{\mu_{#1}(#2)}
\nc{\came}[3]{\mu^\dag_{#1}(#2,#3)}
\nc{\cme}[2]{\mu'_{#1}(#2)}
\nc{\asi}[2]{\sib{(#1,#2)}}
\nc{\ch}{c}
\nc{\cht}{d}
\nc{\chs}{\mathcal{C}}
\nc{\ami}{\arc'}
\nc{\vmi}{\ver'}
\nc{\asd}{s^\dag}
\nc{\camf}{m^\dag}
\nc{\ain}{x}
\nc{\cfs}{\mathcal{E}}
\nc{\cf}{\sigma}
\nc{\lopf}{\hat{y}}
\nc{\ainp}{\epsilon}
\nc{\sop}{\epsilon}
\nc{\cex}[2]{\mathbb{E}[#1\,|\,#2]}
\nc{\arlp}{\arl}
\nc{\gam}[2]{\gamma(#1,#2)}
\nc{\opp}{y'}
\nc{\mtu}{\mathcal{M}}
\nc{\pif}{\pi}
\nc{\ann}{c}
\nc{\ashd}{\ash^\dag}
\nc{\qm}{q}
\nc{\qmp}{\qm'}
\nc{\dep}{d}
\nc{\delg}[3]{\delta_{#1}(#2,#3)}
\nc{\sm}{SM}
\nc{\dm}{DM}
\nc{\aar}{a}
\nc{\hei}{d}
\nc{\pata}[2]{\vartheta_{#1}(#2)}
\nc{\patb}[2]{\vartheta'_{#1}(#2)}
\nc{\patc}[2]{\vartheta^\dag_{#1}(#2)}
\nc{\cvsd}[1]{\cvs_{#1}}
\nc{\cvsdl}[1]{\cvs_{#1}^{\triangleleft}}
\nc{\cvsdr}[1]{\cvs_{#1}^{\triangleright}}
\nc{\dx}{i}
\nc{\alga}{\textsc{Xprop$^*$}}
\nc{\pus}[1]{\textsc{Update}(#1)}
\nc{\cus}[1]{\textsc{Update}'(#1)}
\nc{\sbt}[1]{\textsc{Subtree}(#1)}
\nc{\trns}{\mathcal{D}}
\nc{\trni}{\mathcal{X}'}
\nc{\apf}{\xi}
\nc{\acf}{\xi'}
\nc{\acaf}{\xi^\dag}

\title{Extraction Propagation}

\author{
  Stephen Pasteris\\
  The Alan Turing Institute\\
  London, UK\\
  \texttt{spasteris@turing.ac.uk}
  \and
  Chris Hicks\\
  The Alan Turing Institute\\
  London, UK\\
  \texttt{c.hicks@turing.ac.uk}\\
  \and
  Vasilios Mavroudis\\
  The Alan Turing Institute\\
  London, UK\\
  \texttt{vmavroudis@turing.ac.uk}
}

\date{}

\begin{document}

\maketitle

\begin{abstract}
Running backpropagation end to end on large neural networks is fraught with difficulties like vanishing gradients and degradation. In this paper we present an alternative architecture composed of many small neural networks that interact with one another. Instead of propagating gradients back through the architecture we propagate vector-valued messages computed via forward passes, which are then used to update the parameters. Currently the performance is conjectured as we are yet to implement the architecture. However, we do back it up with some theory. A previous version of this paper was entitled "Fusion encoder networks" and detailed a slightly different architecture.
\end{abstract}

\section{Introduction}

Running backpropagation \cite{Rumelhart1986LearningRB} end to end on large neural networks is fraught with difficulties like vanishing gradients \cite{Hochreiter2001GradientFI} and degradation \cite{He2015DeepRL}. In order to resolve these issues we propose a novel neural network architecture called an \emp{extraction network} (a.k.a. exnet), although we note that the performance is currently only conjectured as we have yet to implement it. An exnet is a directed acyclic graph where each vertex and arc are associated with small neural networks. Whilst some of these neural networks are involved in computing predictions (or any other type of output), many are used solely for updating the parameters (including their own). Our algorithm \emp{Extraction propagation} (a.k.a. \alg) uses the component neural networks to propagate vector-valued messages called \emp{extractions} around the graph. The extractions are then used to construct the output/prediction and update the parameters of the component neural networks. Although the component neural networks are each updated with backpropagation (which should work well due to the small size of these neural networks), the extractions that seed these backpropagations are all computed by forward-passes. Specifically, when the exnet graph is a tree, the scheduling of the message propagation is as in \emp{Belief propagation} \cite{Pearl1982ReverendBO}, where a message propagates in each direction across each edge, and the message propagating from a vertex to its neighbour is constructed from the incoming messages from its other neighbours. In \alg\ this message is constructed (from the incoming messages) via a neural network called a \emp{propagator}. The idea behind \alg\ is that the architecture learns so that the pair of messages propagating across an edge can be used to construct the correct output/prediction, via a neural network called a \emp{trainer}. On each edge, the pair of propagators (one for each direction), coupled with the trainer, forms the neural network to be updated via backpropagation. We note that this is only for the case in which the exnet is a tree: generic exnets generalise this process. Although the performance of \alg\ is currently only conjectured, we do back it up with some theory (based on the assumption that the component neural networks converge optimally). We note that other methods have been proposed for tackling the issues of vanishing gradients and degradation, such as \emp{Long short term memory} \cite{Hochreiter1997LongSM}, \emp{Batch normalisation} \cite{Ioffe2015BatchNA} and \emp{Residual neural networks} \cite{He2015DeepRL}. We hope that \alg\ will be competitive.

We note that a previous version of this paper described a slightly different architecture that we called a \emp{fusion encoder network} and was designed for sequential tasks only (being similar to a tree-structured exnet). If interested in fusion encoder networks please see version 2 of this paper. We also note that version 3 of this paper described only tree-structured exnets and is hence perhaps easier to read.

We now describe the structure of the paper. In Section \ref{defsec} we make the required definitions. In Section \ref{probsec} we formally introduce the problem to be solved. In Section \ref{archsec} we introduce the architecture/algorithm. In Section \ref{ansec} we analyse \alg, showing the intuition behind it. In Section \ref{examsec} we give example exnets: specifically, tree structured exnets for simple sequence/image processing, multi-layered exnets for unstructured instances, and exnets which have an inherent \emp{attention} \cite{Vaswani2017AttentionIA} mechanism for complex tasks. Finally, in Section \ref{altasec} we modify \alg\ to give an alternative algorithm \alga\ that is designed to eradicate a potential issue in \alg.

\section{Definitions}\label{defsec}

Given some real value $a$ that is dependent on an euclidean vector $b$ we denote by $\der{b}a$ the derivative of $a$ with respect to $b$. We let $\mathbb{N}$ be the set of natural numbers excluding $0$. Given $a\in\mathbb{N}$, we define $[a]$ to be the set of all $b\in\mathbb{N}$ with $b\leq a$.

A \emp{directed graph} is a pair of sets $(\ver,\arc)$ where $\arc\subseteq\ver\times\ver$. We call the elements of $\ver$ and $\arc$ \emp{vertices} and \emp{arcs} respectively. A \emp{root} is any vertex $u\in\ver$ such that there does not exist any $v\in\ver$ with $(v,u)\in\arc$. A \emp{leaf} is any vertex $u\in\ver$ such that there does not exist any $v\in\ver$ with $(u,v)\in\arc$. An \emp{internal vertex} is any vertex which is not a leaf. Given vertices $u,v\in\ver$, we say that $v$ is a \emp{child} of $u$ iff $(u,v)\in\arc$. Conversly, $u$ is a \emp{parent} of $v$ iff $v$ is a child of $u$.  A \emp{cycle} is any sequence of vertices $\seq{u_i}{i\in [\arnn]}$ (for any $\arnn\in\nat$) such that $(u_\arnn, u_1)\in\arc$ and, for all $i\in[\arnn-1]$, we have $(u_i, u_{i+1})\in\arc$\,. A \emp{directed acyclic graph} (DAG) is any directed graph with no cycle.

\section{Problem Description}\label{probsec}

We have some set  $\ins$ of \emp{instances} and some euclidean space $\ops$ of \emp{predictions}. Let $\lss$ be the set of all differentiable functions that map $\ops$ into $\real$. We call the elements of $\lss$ \emp{loss functions}. We assume that there exists some unknown probability distribution $\pd$ over $\ins\times\lss$. Learning proceeds in trials where on each trial $t$:
\begin{enumerate}
\item A pair $(\nint{t},\lost{t})$ is drawn from $\pd$.
\item The instance $\nint{t}$ is revealed to us.
\item We must choose some prediction $\nopt{t}\in\ops$
\item The loss function $\lost{t}$ is revealed to us.
\end{enumerate}
Our aim is to learn so that, eventually, we have that $\nopt{t}$ approximately minimises $\cexpt{\lost{t}}{\nint{t}}$.

\section{The Architecture}\label{archsec}

We now describe the extraction network architecture and the Extraction propagation algorithm.

\subsection{Neural Networks}

We first introduce the three fundamental neural networks in our architecture. We have a pair of euclidean spaces $\pss$ and $\css$. We call the elements of $\pss$ \emp{primary extractions} and the elements of $\css$ \emp{complementary extractions}. We also have euclidean spaces $\pp$, $\cp$ and $\tp$ which will parameterise our neural networks. We define the \emp{primary propagator}:
\be
\ppr:\pp\times\pss\times\pss\rightarrow\pss
\ee
the \emp{complementary propagator}:
\be
\cpr:\cp\times\css\times\pss\rightarrow\css
\ee
and the \emp{trainer}:
\be
\trr:\tp\times\pss\times\css\rightarrow\ops
\ee
These functions are neural network structures. The first argument in each function is the parameterisation of the neural network, and the remaining two arguments are concatenated to make the input to the neural network. For example, a primary propagator neural network is defined by a parameterisation $\app\in\pp$, and when vectors $\aps,\acs\in\pss$ are concatenated and inputted into the neural network, the output of the neural network is equal to $\ppr(\app,\aps, \acs)$. Note that a neural network is defined by a parameterisation (in $\pp$, $\cp$ or $\tp$ for the primary propagator, complementary propagator and trainer respectively).

\subsection{Extraction Networks}

Our architecture is called an \emp{extraction network} (a.k.a. \emp{exnet}). An exnet is defined by a DAG $(\ver, \arc)$ with the following properties:
\begin{itemize}
\item There exists a single root.
\item Each internal vertex has exactly two children.
\end{itemize}
Let $\rot$ be the unique root and let $\ips$ be the set of leaves. The exnet is accompanied by a function $\tokf:\ins\rightarrow\pss^\ips$ called the \emp{tokeniser}. Given any internal vertex $z\in\ver\setminus\ips$ we call one of its children its \emp{left-child} and the other its \emp{right-child}, which are denoted as $\lch{z}$ and $\rch{z}$ respectively. Given any arc $(z,v)\in\arc$ we define $\asi{z}{v}$ as follows:
\begin{itemize}
\item If $v=\lch{z}$ then  $\asi{z}{v}:=\rch{z}$
\item If $v=\rch{z}$ then $\asi{z}{v}:=\lch{z}$
\end{itemize}
We call $\asi{z}{v}$ the \emp{sibling} of $v$ with respect to $z$. Given any vertex $v\in\ver$, we denote the set of all its parents by $\pars{v}$. On each trial $t$ the exnet contains the following (parameterisations of) neural networks:
\begin{itemize}
\item For every $v\in\ver\setminus\ips$ we have some $\ppi{t}{v}\in\pp$ and $\tpi{t}{v}\in\tp$.
\item For every $(z,v)\in\arc$ we have some $\cpi{t}{z}{v}\in\cp$
\end{itemize}

Note that we can transform any directed acyclic graph $(\ver,\arc)$ with a single root into an exnet as follows. First, suppose we have a vertex $z\in\ver$ with a single child $v$. Then, given that $\chs$ is the set of children of $v$, remove $v$ from $\ver$, remove $(z,v)$ from $\arc$, and for all $\ch\in\chs$ add $(z,\ch)$ to $\arc$. By repeated performing this operation we remove all vertices with a single child. Next, suppose we have a vertex $z\in\ver$ with more than two children. Then, letting $\chs$ be the set of children of $z$, partition $\chs$ into sets (with cardinalities differing by no more than one) $\chs_1$ and $\chs_2$. Now add two new vertices $v_1$ and $v_2$ to $\ver$. Add $(z,v_1)$ and $(z,v_2)$ to $\arc$ and for all $\ch\in\chs$ remove $(z,\ch)$ from $\arc$. Finally, for all $\ch\in\chs_1$ add $(v_1,\ch)$ to $\arc$ and for all $\ch\in\chs_2$ add $(v_2,\ch)$ to $\arc$. By repeatedly performing this operation we ensure that each internal vertex has exactly two children.

\subsection{Extraction Propagation}
We now describe our algorithm \alg\ (e{\bf{X}}traction {\bf{prop}}agation) which works on an exnet $(\ver, \arc)$, and on any trial $t$, computes $\nopt{t}$ and updates the parameterisations of the neural networks in the exnet. For simplicity, we will describe \alg\ using simple stochastic gradient descent as our optimiser, which requires some \emp{learning rate} $\lr>0$. \alg\ can run in one of two modes: \emp{stochastic mode} or \emp{deterministic mode}. In the following pseudocode, any bullet points labeled \sm\ are only included when in stochastic mode. Similarly, any bullet points labelled \dm\ are only included when in deterministic mode. On trial $t$, \alg\ does the following:
\begin{enumerate}
\item \label{ep1} Define $\inst{t}:=\tokf(\nint{t})$.
\item \label{ep2} For all $v\in\ips$ define $\pme{t}{v}:=\inst{t}(v)$
\item \label{ep3} For all $v\in\ver\setminus\ips$, once $\pme{t}{\lch{v}}$ and $\pme{t}{\rch{v}}$ have been constructed, define:
\be
\pme{t}{v}:=\ppr(\ppi{t}{v}, \pme{t}{\lch{v}}, \pme{t}{\rch{v}})
\ee
\item \label{ep4} Predict $\nopt{t}:=\trr(\tpi{t}{\rot}, \pme{t}{\rot}, 0)$
\item \label{ep5} Receive $\lost{t}$
\item \label{ep6} Define $\cme{t}{\rot}:=0$
\item \label{ep7} For all $v\in\ver\setminus(\ips\cup\{\rot\})$, once $\cme{t}{z}$ has been computed for all $z\in\pars{v}$ we do as follows:
\begin{enumerate}
\item \label{ep7a} For all $z\in\pars{v}$ define:
\be
\came{t}{z}{v} := \cpr(\cpi{t}{z}{v},\cme{t}{z}, \pme{t}{\asi{z}{v}})
\ee
\item \label{ep7b} 
\begin{itemize}
\item SM. Select $z$ uniformly at random from $\pars{v}$ and define: 
\be
\cme{t}{v} := \came{t}{z}{v}
\ee
\item DM. Define:
\be
\cme{t}{v} := \sum_{z\in\pars{v}}\came{t}{z}{v}
\ee
\end{itemize}
\end{enumerate}
\item \label{ep8} For all $v\in\ver\setminus\ips$ define:
\be
\tpi{t+1}{v}:=\tpi{t}{v}-\lr\der{\tpi{t}{v}}\lost{t}(\trr(\tpi{t}{v}, \pme{t}{v}, \cme{t}{v}))
\ee
\be
\ppi{t+1}{v}:=\ppi{t}{v}-\lr\der{\ppi{t}{v}}\lost{t}(\trr(\tpi{t}{v}, \ppr(\ppi{t}{v}, \pme{t}{\lch{v}}, \pme{t}{\rch{v}}), \cme{t}{v}))
\ee
\item \label{ep9} For all $(z,v)\in\arc$ with $v\notin\ips$, define:
\begin{itemize}
\item SM:
\be
\delg{t}{z}{v}:=\der{\cpi{t}{z}{v}}\lost{t}(\trr(\tpi{t}{v}, \pme{t}{v}, \cpr(\cpi{t}{z}{v}, \cme{t}{z}, \pme{t}{\asi{z}{v}}))
\ee
\item DM:
\be
\delg{t}{z}{v}:=\der{\cpi{t}{z}{v}}\lost{t}\left(\trr\left(\tpi{t}{v}, \pme{t}{v}, \sum_{\hat{z}\in\pars{v}}\cpr(\cpi{t}{\hat{z}}{v}, \cme{t}{\hat{z}}, \pme{t}{\asi{\hat{z}}{v}}\right)\right)
\ee
\end{itemize}
and then define:
\be
\cpi{t+1}{z}{v}:=\cpi{t}{z}{v}-\lr\delg{t}{z}{v}
\ee
\end{enumerate}

We note that, in the above, our parameter updates are simple gradient descent updates. Typically we would want to use, instead, a more refined optimiser such as ADAM \cite{8624183} (which will utilise the same gradients as in the pseudocode). It is also possible to, as in convolutional neural networks \cite{Fukushima1980NeocognitronAS, Li2020ASO}, share the parameters of neural networks on different vertices/arcs. If parameters are shared across multiple vertices/arcs then the gradients with respect to each are computed and summed together, before subtraction from the current parameterisation.

\subsection{Description}

We now describe the \alg\ algorithm given in the above pseudocode. \alg\ consists of three phases: an \emp{up pass} (lines \ref{ep1} to \ref{ep4}) in which primary extractions are computed recursively from the leaves to the root and the prediction $\nopt{t}$ is generated, a \emp{down pass} (lines \ref{ep6} to \ref{ep7}) in which, using the previously computed primary extractions, complementary extractions are computed recursively from the root to the leaves, and a \emp{parameter update} (line \ref{ep8} to \ref{ep9}) in which, using the previously computed primary and complementary extractions, a gradient descent step is taken for all the neural network parameters.

\begin{figure}[h]
    \centering
    \includegraphics[width=0.8\textwidth]{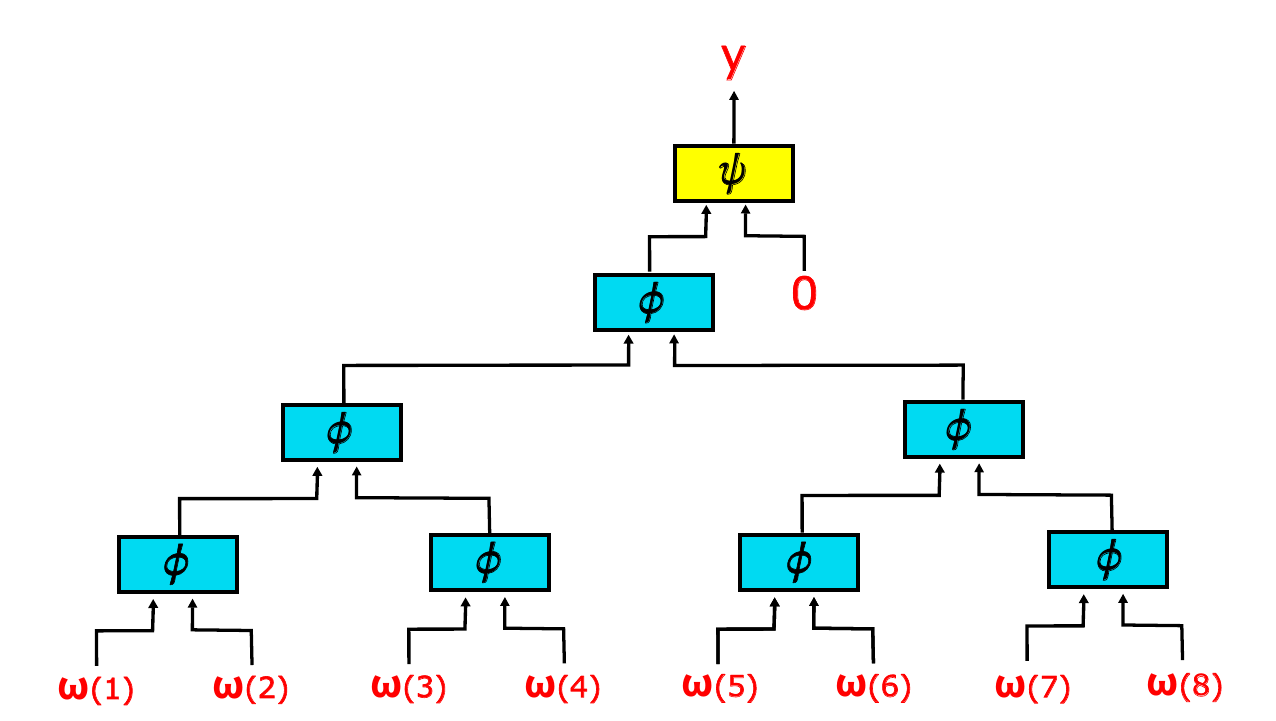}
    \caption{The primary architecture when $\ips=[8]$ and the exnet is a balanced tree. The subscript of $t$ has been dropped from all vectors.}
    \label{priarcfig}
\end{figure}

We first describe the up pass. Firstly the instance $\nint{t}$ is \emp{tokenised} which involves the creation of a primary extraction $\inst{t}(v)\in\pss$ for each leaf $v\in\ips$ in some arbitrary way (that is consistent from trial to trial). If the instance is an euclidean vector this could involve simply partitioning the components of $\nint{t}$ into collections (one for each $v\in\ips$) each with cardinality equal to the dimensionality of $\pss$, and then converting each collection into the vector $\inst{t}(v)$. After the tokenisation, primary extractions are propagated up the exnet from the leaves to the root. Specifically, for each vertex $v\in\ver$ a primary extraction $\pme{t}{v}\in\pss$ is constructed as follows. For each leaf $v\in\ips$ we set $\pme{t}{v}:=\inst{t}(v)$. For each internal vertex $v\in\ver\setminus\ips$, once we have constructed the primary extractions for both children of $v$, the primary extraction $\pme{t}{v}$ is constructed from $\pme{t}{\lch{v}}$ and $\pme{t}{\rch{v}}$ by the primary propagator neural network at vertex $v$. After the construction of all the primary extractions, the trainer neural network at the root $\rot$ then computes the prediction $\nopt{t}$ from $\pme{t}{\rot}$ and the zero vector in $\css$. We note that the prediction $\nopt{t}$ is actually just the output of a large neural network, which we call the \emp{primary architecture}, that is formed from all the primary propagator neural networks on the internal vertices and the trainer neural network on $\rot$. The primary architecture of a tree-structured exnet is shown in Figure \ref{priarcfig}. The novelty of \alg, however, is how this large neural network is updated. In order to update it is important for us to cache (for the next two phases of \alg) all the primary extractions created in the up pass.

We now describe the down pass, in which complementary extractions are propagated down the exnet from the root to the leaves. Specifically, for each internal vertex $v\in\ver\setminus\ips$ a complementary extraction $\cme{t}{v}\in\css$ is constructed, and for each arc $(z,v)$ with $v\notin\ips$ a complementary extraction $\came{t}{z}{v}\in\css$ is constructed. First, the extraction $\cme{t}{\rot}$ is defined equal to the zero vector and hence contains no information. For any vertex $v\in\ver\setminus(\ips\cup\{\rot\})$, once the complementary extraction has been computed for each parent of $v$, the following extractions are then computed. For each parent $z\in\pars{v}$ the extraction $\came{t}{z}{v}$ is computed, via the complementary propagator neural network on the arc $(z,v)$, from the complementary extraction $\cme{t}{z}$ and the primary extraction (computed in the up pass) for the sibling $\asi{z}{v}$. When in stochastic mode, the extraction $\cme{t}{z}$ is then chosen uniformly at random from these extractions (i.e. $z$ is chosen uniformly at random from $\pars{v}$ and then $\cme{t}{v}$ is set equal to $\came{t}{z}{v}$). When in deterministic mode, these extractions are instead summed to create $\cme{t}{v}$.

\begin{figure}[h]
    \centering
    \includegraphics[width=0.8\textwidth]{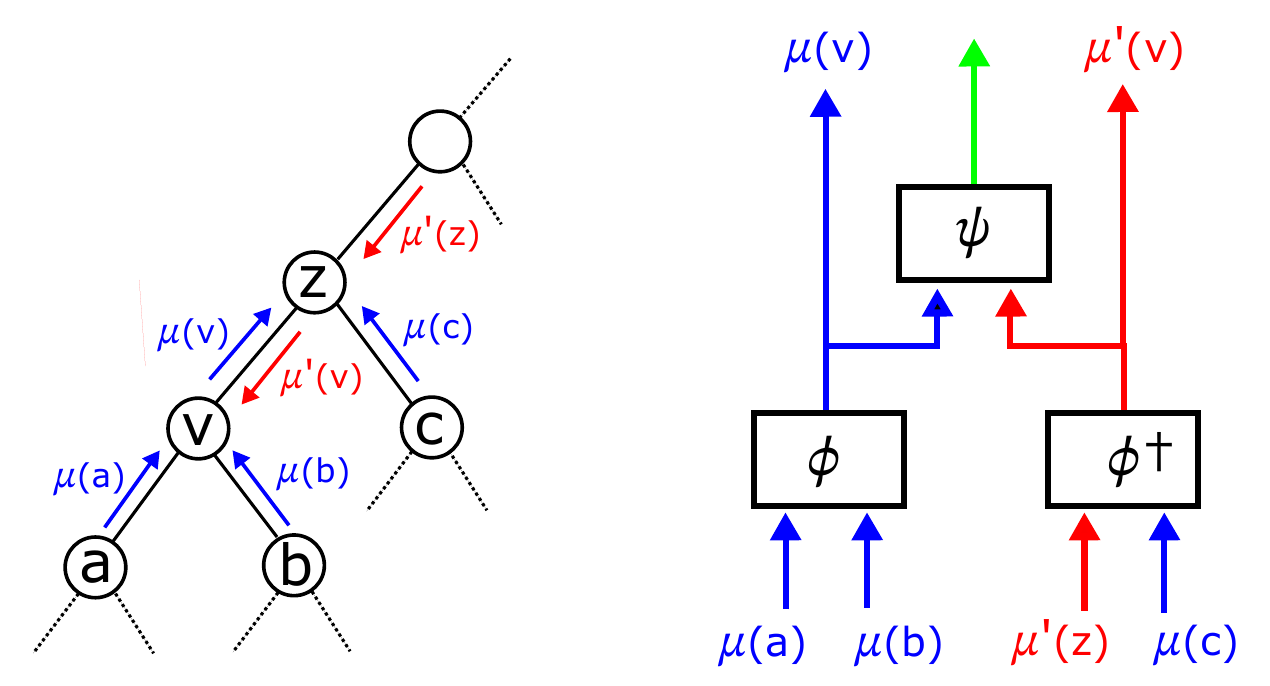}
    \caption{Extraction computation and parameter updates at a vertex $v$ with a single parent $z$, noting that $\cme{t}{v}=\came{t}{z}{v}$. The subscript $t$ has been dropped from all extractions. The left hand side depicts the vertices and extractions involved. The right hand side depicts the neural networks involved as well as how the extractions are computed. The neural networks are updated by backpropagation from the (gradient of the loss of the) prediction denoted by the green arrow. Note that blue and red indicate primary and complementary extractions respectively.}
    \label{extfig}
\end{figure}

We now describe the parameter update. The idea is that we want to learn so that, eventually, for each internal vertex $v\in\ver\setminus\ips$ we have that $\pme{t}{v}$ and $\cme{t}{v}$ contain, between them, enough information to compute an approximate minimiser of $\cexpt{\lost{t}}{\nint{t}}$. The purpose of the trainer neural network at vertex $v$ is to compute this approximate minimiser. In other words, we want the value:
\begin{equation}\label{parupeq1}
\lost{t}(\trr(\tpi{t}{v}, \pme{t}{v}, \cme{t}{v}))
\end{equation}
to be approximately minimised in expectation. Note then that the fact that the prediction $\nopt{t}$ is computed at the root $\rot$ is arbitrary: it can, in fact, be computed at any internal vertex. Equation \eqref{parupeq1} can be re-written as:
\begin{equation}\label{parupeq2}
\lost{t}(\trr(\tpi{t}{v}, \ppr(\ppi{t}{v}, \pme{t}{\lch{v}}, \pme{t}{\rch{v}}), \cme{t}{v}))
\end{equation}
Since we want the (equivalent) terms in equations \eqref{parupeq1} and \eqref{parupeq2} to be minimised in expectation, we update the parameters of our primary propagators and trainers via a gradient descent step, taking the gradient with respect to $\tpi{t}{v}$ in Equation \eqref{parupeq1} and the gradient with respect to $\ppi{t}{v}$ in Equation \ref{parupeq2}. We now turn to the updates of our complementary propagators, which differ depending on the mode. First consider stochastic mode. Recall that in stochastic mode $\cme{t}{v}$ is equal to $\came{t}{z}{v}$ for some random $z\in\pars{v}$, meaning Equation \ref{parupeq1} can be re-written as:
\be
\lost{t}(\trr(\tpi{t}{v}, \pme{t}{v}, \cpr(\cpi{t}{z}{v}, \cme{t}{z}, \pme{t}{\asi{z}{v}})))
\ee
Hence, for all $z\in\pars{v}$ we update $\cpi{t}{z}{v}$ via a gradient descent step with respect to this term. Now consider deterministic mode. Here we can rewrite Equation \ref{parupeq1} as:
\be
\lost{t}\left(\trr\left(\tpi{t}{v}, \pme{t}{v}, \sum_{\hat{z}\in\pars{v}}\cpr(\cpi{t}{\hat{z}}{v}, \cme{t}{\hat{z}}, \pme{t}{\asi{\hat{z}}{v}})\right)\right)
\ee
so, for all $z\in\pars{v}$ we update $\cpi{t}{z}{v}$ via a gradient descent step with respect to this term.  Figure \ref{extfig} depicts the computation of the extractions and the parameter updates when the exnet is a tree (noting that in this case the two modes are identical to each other).

\section{Analysis}\label{ansec}

In this section we give the intuition behind our architecture: analysing the eventual result of \alg, in stochastic mode, under the assumption that the combinations of the neural networks at each vertex converge to their optima. We show that, importantly, such convergence is not as far-fetched as it might first appear. We also show the crucial result that information learnt in one part of the exnet is propagated to the rest of the exnet, and hence incorporated in the prediction.

We note that whilst, in this analysis, we consider \alg\ in stochastic mode, the argument immediately carries over into deterministic mode. However, the intuition is clearer when studying stochastic mode which is why we have chosen this mode for the analysis.

\subsection{Our Functions in the Limit}

First define $\vmi:=\ver\setminus\ips$ and define $\ami$ to be the set of all arcs $(z,v)\in\arc$ with $v\notin\ips$. Assuming convergence, we define the functions:
\be
\pff:\vmi\times\pss\times\pss\rightarrow\pss~~~~~;~~~~~\cff:\ami\times\css\times\pss\rightarrow\css~~~~~;~~~~~\gf:\vmi\times\pss\times\css\rightarrow\ops
\ee
such that for all $v\in\vmi$ and $\as,\ash\in\pss$ we have:
\be
\pff(v,\as,\ash):=\lim_{t\rightarrow\infty}\ppr(\ppi{t}{v}, \as, \ash)
\ee
and for all $(z,v)\in\ami$, $\asd\in\css$ and $\as\in\pss$ we have:
\be
\cff((z,v),\asd,\as):=\lim_{t\rightarrow\infty}\cpr(\cpi{t}{z}{v}, \asd, \as)
\ee
and for all $v\in\vmi$, $\as\in\pss$ and $\asd\in\css$ we have:
\be
\gf(v,\as,\asd):=\lim_{t\rightarrow\infty}\trr(\tpi{t}{v}, \as, \asd)
\ee
We also define the \emp{choice set} $\cfs$ as the set of functions $\cf$ with domain $\vmi\setminus\{\rot\}$ such that for all $v\in\vmi\setminus\{\rot\}$ we have:
\be
\cf(v)\in\pars{v}
\ee
We now recursively construct the functions:
\be
\pmf:\ver\times\ins\rightarrow\pss~~~~~;~~~~~\cmf:\vmi\times\ins\times\cfs\rightarrow\css~~~~~;~~~~~\camf:\ami\times\ins\times\cfs\rightarrow\css
\ee
as follows. For all $v\in\ips$ and $\ain\in\ins$ we define:
\be
\pmf(v,\ain):=\tokf(\ain)(v)
\ee
and for all $v\in\vmi$ and $\ain\in\ins$ we recursively define:
\be
\pmf(v,\ain):=\pff(v,\pmf(\lch{v},\ain),\pmf(\rch{v},\ain))
\ee
Now that we have the function $\pmf$, we define the functions $\cmf$ and $\camf$ as follows. For all $\ain\in\ins$ and $\cf\in\cfs$ we define:
\be
\cmf(\rot,\ain,\cf):=0
\ee
For all $v\in\vmi\setminus\{\rot\}$, once $\cmf(z,\blank,\blank)$ has been defined for all $z\in\pars{v}$ we first define, for all $z\in\pars{v}$, $\ain\in\ins$ and $\cf\in\cfs$, the quantity:
\be
\camf((z,v),\ain,\cf):=\cff((z,v),\cmf(z,\ain,\cf),\pmf(\asi{z}{v},\ain))
\ee
and we then define, for all $\ain\in\ins$ and $\cf\in\cfs$, the quantity:
\be
\cmf(v,\ain,\cf):=\camf((\cf(v),v),\ain,\cf)
\ee
Finally, we define the \emp{local prediction} function:
\be
\lopf:\vmi\times\ins\times\cfs\rightarrow\ops
\ee
such that for all $v\in\vmi$, $\ain\in\ins$ and $\cf\in\cfs$ we have:
\be
\lopf(v,\ain,\cf):=\gf(v,\pmf(v,\ain),\cmf(v,\ain,\cf))
\ee
It will become clear in the next subsection that in the limit $t\rightarrow\infty$ we have that:
\be
\nopt{t}=\lopf(\rot,\nint{t},\cf)
\ee
noting that this value is independent of $\cf$.

\subsection{Vertex Optimality}

For all $v\in\vmi\setminus\{\rot\}$, $\ain\in\ins$ and $\cf\in\cfs$ we define the tuple:
\be
\mtu(v,\ain,\cf):=(\pmf(\lch{v},\ain)\,, \pmf(\rch{v},\ain)\,, \cmf(\cf(v),\ain,\cf)\,, \pmf(\asi{\cf(v)}{v},\ain))
\ee
We also define, for all $\ain\in\ins$ and $\cf\in\cfs$, the tuple:
\be
\mtu(\rot,\ain,\cf):=(\pmf(\lch{\rot},\ain), \pmf(\rch{\rot},\ain), 0, 0)
\ee
In what follows, given $(z,v)\in\ami$ and $\cf\in\cfs$ we will define $\ainp$, $\arlp$ and $\gam{z}{v}$ to be random variables such that $(\ainp,\arlp)\in\ins\times\lss$ is drawn from $\pd$ and, independently, $\gam{z}{v}$ is drawn uniformly at random from the set of all $\cf\in\cfs$ with $\cf(v)=z$.

Note that, upon convergence, \alg\ (in stochastic mode) behaves as follows. As $t\rightarrow\infty$ we have, for all $v\in\ver$, that:
\be
\pme{t}{v}=\pmf(v,\nint{t})
\ee
and then some $\cf$ is drawn uniformly at random from $\cfs$ and we have, for all arcs $(z,v)\in\ami$ that:
\be
\came{t}{z}{v}=\camf((z,v),\nint{t},\cf)~~~~~;~~~~~\cme{t}{v}=\cmf(v,\nint{t},\cf)
\ee
and hence, for all $v\in\vmi$, we have that $\lopf(v,\nint{t},\cf)$ is the output of the trainer at vertex $v$.
Hence, assuming that the aggregated neural networks at each vertex/arc converge to their optima, we will have, for all $v\in\vmi$, $\ain\in\ins$ and $\cf\in\cfs$, that:
\begin{equation}\label{lopeq1}
\lopf(v,\ain,\cf)=\operatorname{argmin}_{\opp\in\ops}\cex{\arlp(\opp)}{\mtu(v,\ainp,\gam{\cf(v)}{v})=\mtu(v,\ain,\cf)}
\end{equation}
We shall call such a property \emp{vertex optimality}. For simplicity, we will assume that the minimiser in Equation \eqref{lopeq1} is unique.

On first glance, it may appear that achieving vertex optimality would require a very large complementary extraction space $\css$ due to the potentially massive cardinality of $\cfs$. However, we will show in the next subsection that, given vertex optimality, $\lopf(v,\ain,\cf)$ is independent of $\cf$. This means that, crucially, by the definition of $\lopf$, we can achieve vertex optimaility with the extraction $\cmf(v,\ain,\cf)$ being independent of $\cf$. Having $\cmf(v,\ain,\cf)$ independent of $\cf$ for all $v\in\vmi$ means that (typically) the space $\css$ can have a relatively low dimension.

\subsection{Identical Local Predictions}

We now prove that, given vertex optimality, we have that, for all $\ain\in\ins$, the function $\cmf(\blank,\ain,\blank)$ is constant. Not only does this imply the above crucial property, but it means that any information learnt in some part of the exnet is propagated to the rest of the exnet.

To prove this, first note that since $\cmf(\rot,\ain,\blank)$ is constant, the function $\lopf(\rot,\ain,\blank)$ is also constant. Hence, define the function $\pif:\ins\rightarrow\ops$ such that for all $\ain\in\ins$ and $\cf\in\cfs$ we have:
\be
\pif(\ain):=\lopf(\rot,\ain,\cf)
\ee
We then take the inductive hypothesis that, for any given $v\in\vmi$, we have, for all $\ain\in\ins$ and $\cf\in\cfs$ that:
\be
\lopf(v,\ain,\cf)=\pif(\ain)
\ee
Since the directed graph $(\ver,\arc)$ is acyclic with a single root, we can prove this via induction from the root to the leaves. By definition of $\pif$ the inductive hypothesis immediately holds for $v=\rot$. Hence, all that we need is to show is that, for any vertex $\ann\in\vmi\setminus\{\rot\}$, if the inductive hypothesis holds for all $v\in\pars{\ann}$ then the inductive hypothesis also holds for $v=\ann$. So suppose we have a vertex $\ann\in\vmi\setminus\{\rot\}$ such that the inductive hypothesis holds for all $v\in\pars{\ann}$. Take any $\cf\in\cfs$ and let $z:=\cf(v)$. Without loss of generality assume that $\ann=\lch{z}$. Define:
\be
\as:=\pmf(\ann,\ain)~~~~~;~~~~~\asp:=\pmf(\asi{z}{\ann},\ain)~~~~~;~~~~~\asd:=\cmf(z,\ain,\cf)
\ee
In the case that $z=\rot$ we have: 
\be
\asd=0=\cmf(z,\ainp,\gam{z}{\ann}))
\ee 
so since $\ann=\lch{z}$ and $\asi{z}{\ann}=\rch{z}$ we immediately have, from Equation \eqref{lopeq1}, that:
\be
\lopf(z,\ain,\cf)=\operatorname{argmin}_{\opp\in\ops}\cex{\arlp(\opp)}{\pmf(\ann,\ainp)=\as\,, \pmf(\asi{z}{\ann},\ainp)=\asp\,, \cmf(z,\ainp,\gam{z}{\ann}))=\asd}
\ee
so, by the inductive hypothesis, we have:
\begin{equation}
\operatorname{argmin}_{\opp\in\ops}\cex{\arlp(\opp)}{\pmf(\ann,\ainp)=\as\,, \pmf(\asi{z}{\ann},\ainp)=\asp\,, \cmf(z,\ainp,\gam{z}{\ann})=\asd}=\pif(\ain)
\end{equation}
In the case that $z\neq\rot$ let: 
\be
\aar:=(\cf(z),z)~~~~~;~~~~~\ashd:=\cmf(\cf(z),\ain, \cf)~~~~~;~~~~~\ash:=\pmf(\sib{\aar},\ain)
\ee
Note that, since $\ann=\lch{z}$ and $\asi{z}{\ann}=\rch{z}$, we have:
\be
\pmf(z,\ain)=\pff(z, \as, \asp)~~~~~;~~~~~\cmf(z,\ain,\cf)=\cff(\aar,\ashd,\ash)
\ee
so that, by definition of $\lopf$, we have:
\be
\lopf(z,\ain,\cf)=\gf(z, \pff(z,\as,\asp), \cff(\aar,\ashd,\ash))
\ee
Noting that:
\be
\mtu(z,\ain,\cf)=(\as,\asp,\ashd,\ash)
\ee
we then have, by Equation \eqref{lopeq1}, that $\lopf(z,\ain,\cf)$ minimises:
\be
\cex{\arlp}{\pmf(\ann,\ainp)=\as\,, \pmf(\asi{z}{\ann},\ainp)=\asp\,, \cff(\aar,\cmf(\cf(z),\ainp,\gam{\cf(z)}{z}),\pmf(\sib{\aar},\ainp))=\cff(\aar,\ashd,\ash)}
\ee
Since:
\be
\cff(\aar,\ashd,\ash)=\cff(\aar,\cmf(\cf(z),\ain, \cf), \pmf(\sib{\aar},\ain))=\cmf(z,\ain,\cf)=\asd
\ee
and:
\be
\cff(\aar,\cmf(\cf(z),\ainp,\gam{z}{\cf}),\pmf(\sib{\aar},\ainp))=\cmf(z,\ainp,\gam{z}{\cf})
\ee
we then have:
\be
\lopf(z,\ain,\cf)=\operatorname{argmin}_{\opp\in\ops}\cex{\arlp(\opp)}{\pmf(\ann,\ainp)=\as\,, \pmf(\asi{z}{\ann},\ainp)=\asp\,, \cmf(z,\ainp,\gam{\cf(z)}{z})=\asd}
\ee
so, by the inductive hypothesis, we have:
\be
\operatorname{argmin}_{\opp\in\ops}\cex{\arlp(\opp)}{\pmf(\ann,\ainp)=\as\,, \pmf(\asi{z}{\ann},\ainp)=\asp\,, \cmf(z,\ainp,\gam{\cf(z)}{z})=\asd}=\pif(\ain)
\ee
Since the right hand side (and hence left hand side) does not depend on $\cf(z)$ and $\cmf(z,\ainp,\cf')$ is independent of $\cf'(\ann)$, we then have:
\begin{equation}\label{aneq1}
\operatorname{argmin}_{\opp\in\ops}\cex{\arlp(\opp)}{\pmf(\ann,\ainp)=\as\,, \pmf(\asi{z}{\ann},\ainp)=\asp\,, \cmf(z,\ainp,\gam{z}{\ann})=\asd}=\pif(\ain)
\end{equation}
We have now proved Equation \eqref{aneq1} in all cases (whether $z=\rot$ or not).
Now define:
\be
\qm:=\pmf(\lch{\ann},\ain)~~~~~;~~~~~\qmp:=\pmf(\rch{\ann},\ain)
\ee
Since:
\be
\pmf(\ann,\ain)=\pff(\ann,\qm,\qmp)~~~~~;~~~~~\cmf(\ann,\ain,\cf)=\cff((z,\ann),\asd,\asp)
\ee
we have, by definition of $\lopf$, that:
\be
\lopf(\ann,\ain,\cf)=\gf(\ann, \pff(\ann,\qm,\qmp), \cff((z,\ann),\asd,\asp))
\ee
Noting that:
\be
\mtu(\ann,\ain,\cf)=(\qm,\qmp,\asd,\asp)
\ee
we then have, by Equation \eqref{lopeq1}, that $\lopf(\ann,\ain,\cf)$ minimises:
\be
\cex{\arlp}{\pff(\pmf(\lch{\ann},\ainp)\,, \pmf(\rch{\ann},\ainp))=\pff(\qm,\qmp)\,, \cmf(z,\ainp,\gam{z}{\ann})=\asd\,, \pmf(\asi{z}{\ann},\ainp)=\asp}
\ee
Since:
\be
\pff(\qm,\qmp)=\pff(\pmf(\lch{\ann},\ain),\pmf(\rch{\ann},\ain))=\pmf(\ann,\ain)=\as
\ee
and:
\be
\pff(\pmf(\lch{\ann},\ainp)\,, \pmf(\rch{\ann},\ainp))=\pmf(\ann,\ainp)
\ee
we then have that:
\be
\lopf(\ann,\ain,\cf)=\operatorname{argmin}_{\opp\in\ops}\cex{\arlp(\opp)}{\pmf(\ann,\ainp)=\as\,, \pmf(\asi{z}{\ann},\ainp)=\asp\,, \cmf(z,\ainp,\gam{z}{\ann}))=\asd}
\ee
By Equation \eqref{aneq1} we have now shown that:
\be
\lopf(\ann,\ain,\cf)=\pif(\ain)
\ee
and hence the inductive hypothesis holds for $v=\ann$. We have hence shown that the inductive hypothesis holds for all $v\in\vmi$ and hence that, for all $\ain\in\ins$, the function $\lopf(\blank,\ain,\blank)$ is constant, as required.

\section{Example Exnets}\label{examsec}
In this section we give example exnets. Specifically, we construct tree-structured exnets for simple sequence/image processing tasks, we construct multi-layered exnets for unstructured instances, and finally we construct exnets which utilise an \emp{attention} mechanism to learn highly complex tasks.

\subsection{Tree-Structured Exnets}\label{tsenexsec}
In this subsection we show how to construct tree-structured exnets for when our instances are sequences or images. It is straightforward to generalise to higher-order tensors. These exnets exploit the fact that when given a contiguous segment of a sequence or image, it is typically easy to compress the relevant information contained within it. We note that the primary extractions correspond to such compressions whilst the complementary extractions correspond to compressions of the complements of such segments.

Our first example is for sequential instances. Here we have some $\sln\in\nat$ and $\ins=\pss^\sln$. We define our exnet $(\ver,\arc)$ to be a balanced binary tree with $\sln$ leaves. For all $\ix\in[\sln]$ we denote by $\ipn{\ix}$ the $\ix$-th leaf from the left. We define the tokeniser $\tokf$ to be such that for all $\ain\in\pss^\sln$ we have, for all $\ix\in[\sln]$, that $\tokf(\ain)(\ipn{\ix})$ is the $\ix$-th component of $\ain$. It will often be the case that we would want to share parameters in the following way. Given any vertex $v\in\ver\setminus\{\rot\}$ let $\pnt{v}$  be the unique parent of $v$. Given $\dep\in\nat$, all vertices $v$ at depth $\dep$ with $v=\lch{(\pnt{v})}$ share the parameters of the neural networks associated with $v$ and $(\pnt{v},v)$. Similarly, all vertices $v$ at depth $\dep$ with $v=\rch{(\pnt{v})}$ share the parameters of the neural networks associated with $v$ and $(\pnt{v},v)$.

Our second example is for when our instances are images. Here we have some $\sln\in\nat$ and our instances correspond to matrices (a.k.a \emp{images}) in $\pss^{\sln\times\sln}$. We have an \emp{overlap level} $\ovl$ with $1/2\leq\ovl<1$. The intuition behind having an overlap level greater than $1/2$ is that the relevant information in a region of an image is easier to compress if you know, in addition to the region, a margin surrounding it. Our exnet $(\ver,\arc)$ is a tree constructed as follows. The vertices in $\ver$ represent rectangular regions of an image. Specifically, each vertex $v\in\ver$ is associated with numbers $\hr{v},\hrp{v},\vr{v},\vrp{v}\in\real$ and, given an image $\bset\in\tks^{\sln\times\sln}$, represents the region:
\be
\{\set{\ix}{\jx}\,|\,\hr{v}\leq\ix\leq\hrp{v}\,,\,\vr{v}\leq\jx\leq\vrp{v}\}
\ee
The root $\rot$ represents an entire image and hence:
\be
\hr{\rot}:=1 \spa \hrp{\rot}:=\sln \spa \vr{\rot}:=1 \spa \vrp{\rot}:=\sln
\ee
The set $\ver\setminus\ips$ is partitioned into two sets $\vea$ and $\veb$ where both children of a vertex in $\vea$ are in $\veb$ and both children of a vertex in $\veb$ are in $\vea$. We have $\rot\in\vea$. Given a vertex in $\vea$, the region of the image that it represents is split horizontally into two (possibly overlapping) regions, which are the regions represented by its children. Specifically, given $v\in\vea$ we have:
\be
\hr{\lch{v}}:=\hr{v} \spa \hrp{\lch{v}}:=\hr{v}+\ovl(\hrp{v}-\hr{v}) \spa \vr{\lch{v}}=\vr{v} \spa \vrp{\lch{v}}=\vrp{v}
\ee
\be
\hr{\rch{v}}:=\hrp{v}-\ovl(\hrp{v}-\hr{v}) \spa \hrp{\rch{v}}:=\hrp{v} \spa \vr{\rch{v}}=\vr{v} \spa \vrp{\rch{v}}=\vrp{v}
\ee

Given a vertex in $\veb$, the region of the image that it represents is split vertically into two (possibly overlapping) regions, which are the regions represented by its children. Specifically, given $v\in\veb$ we have:
\be
\hr{\lch{v}}:=\hr{v} \spa \hrp{\lch{v}}:=\hrp{v} \spa \vr{\lch{v}}=\vr{v} \spa \vrp{\lch{v}}:=\vr{v}+\ovl(\vrp{v}-\vr{v})
\ee
\be
\hr{\rch{v}}:=\hr{v} \spa \hrp{\rch{v}}:=\hrp{v} \spa \vr{\rch{v}}:=\vrp{v}-\ovl(\vrp{v}-\vr{v}) \spa \vrp{\rch{v}}=\vrp{v}
\ee
If a vertex $v$ is such that $\hrp{v}-\hr{v}<1$ and $\vrp{v}-\vr{v}<1$ then $v\in\ips$ and is hence a leaf. Note then that each leaf corresponds to at most one component in the image. The tokeniser $\tokf$ is defined accordingly (any leaves not corresponding to a component can effectively be ignored by assigning them some \emp{null} primary extraction). Note that we can share parameters as in the above sequential example.

\nc{\nml}{\Lambda}
\nc{\lay}[1]{\mathcal{G}_{#1}}

\subsection{Multi-Layer Exnets}

The following family of exnets is inspired by the classic \emp{multi-layer perceptron} \cite{Rosenblatt1958ThePA, Rumelhart1986LearningRB} and is designed for unstructured instances. We have $\nml$ \emph{layers}, each comprising of a set of vertices. Specifically, for each layer $i\in[\nml]$ we have some $\lay{i}\subseteq\ver$. The layers are pairwise disjoint in that, for all $i,j\in[\nml]$ with $i\neq j$, we have that $\lay{i}\cap\lay{j}=\emptyset$. We have $\lay{1}:=\ips$ and $\lay{\nml}:=\{\rot\}$. 

We now describe the additional vertices in $\ver$ as well as the set of arcs $\arc$. Specifically, for all $i\in[\nml-1]$ and all $v\in\lay{i+1}$ we add vertices and arcs so that we have (as a subgraph of $(\ver,\arc)$) a balanced binary tree rooted at $v$ and with set of leaves $\lay{i}$. Note that all vertices in this tree, except for the root $v$ and the leaves (in $\lay{i}$), are new vertices.

\nc{\posen}[1]{\epsilon_{#1}}
\nc{\lav}[2]{q_{#1,#2}}
\nc{\lavp}[2]{q'_{#1,#2}}
\nc{\dob}[3]{s_{#1,#2,#3}}
\nc{\tree}[2]{\mathcal{T}_{#1,#2}}
\nc{\nmh}{H}
\nc{\hed}{h}
\nc{\tret}[3]{\mathcal{T}_{#1,#2,#3}}
\nc{\htr}[2]{\mathcal{T}'_{#1,#2}}
\nc{\wid}{W}
\nc{\spn}{\mathcal{Z}}
\nc{\snto}[2]{\mathcal{B}(#1,#2)}
\nc{\sntt}[1]{\mathcal{B}'(#1)}
\nc{\fsn}[2]{\zeta(#1,#2)}

\subsection{Attention-Based Exnets}

We now introduce a family of exnets inspired by the \emph{transformer} \cite{Vaswani2017AttentionIA} architecture. Here we have some $\sln\in\nat$ and $\ins=\pss^\sln$. We assume that for all $\ain\in\ins$, each component of $\ain$ contains a \emp{positional encoding} as in a transformer. We have $\sln$ leaves where, for all $\jx\in[\sln]$, we denote by $\ipn{\jx}$ the $\jx$-th leaf. Our tokeniser $\tokf$ is such that for all $\ain\in\ins$ and $\jx\in[\sln]$ we have that $\tokf(\ain)(\ipn{\jx})$ is the $\jx$-th component of $\ain$.

We have $\nml$ \emph{layers} where each layer $\ix\in[\nml]$ is a sequence of vertices $\seq{\lav{\ix}{\jx}}{\jx\in[\sln]}$. The first layer is defined so that for all $\jx\in[\sln]$ we have $\lav{1}{\jx}=\ipn{\jx}$. For each layer $\ix\in[\nml-1]$ we will add vertices/arcs as follows (NB: sometimes, for simplicity of explanation, we will create a vertex whose two children are in fact the same vertex, but it is straightforward to fix this).

First, for all $\jx\in[\sln]$, we create a vertex $\lavp{\ix}{\jx}$ and set $\lch{(\lavp{\ix}{\jx})}:=\lav{\ix}{\jx}$ and $\rch{(\lavp{\ix}{\jx})}:=\ipn{\jx}$. Creating this vertex allows us to mix the original instance back into the layer so as to regain any information that has been lost. It may, however, not be required: meaning that, instead, we can define $\lavp{\ix}{\jx}=\lav{\ix}{\jx}$. Next, for all $\jx,\kx\in[\sln]$ we create a vertex $\dob{\ix}{\jx}{\kx}$ and set $\lch{\dob{\ix}{\jx}{\kx}}:=\lavp{\ix}{\jx}$ and $\rch{\dob{\ix}{\jx}{\kx}}:=\lavp{\ix}{\kx}$. Finally, for all $\jx\in[\sln]$ we add vertices to create (as a subgraph of $(\ver,\arc)$) a balanced binary tree $\tree{\ix}{\jx}$, whose root is $\lav{\ix+1}{\jx}$ and whose leaves are the vertices $\dob{\ix}{\jx}{\kx}$ for all $\kx\in[\sln]$. We note that, apart from the leaves and the root, all the vertices in $\tree{\ix}{\jx}$ are new.

For each layer $\ix\in[\nml-1]$ we share parameters as follows (noting that no two layers share any parameters). For all $\jx\in[\sln]$ the neural networks on the vertices $\lavp{\ix}{\jx}$ share their parameters, the neural networks on the arcs $(\lavp{\ix}{\jx},\lav{\ix}{\jx})$ share their parameters, and the neural networks on the arcs $(\lavp{\ix}{\jx},\ipn{\jx})$ share their parameters. For all $\jx,\kx\in[\sln]$, the neural networks on the vertices $\dob{\ix}{\jx}{\kx}$ share their parameters, the neural networks on the arcs $(\dob{\ix}{\jx}{\kx},\lavp{\ix}{\jx})$ share their parameters, and the neural networks on the arcs $(\dob{\ix}{\jx}{\kx},\lavp{\ix}{\kx})$ share their parameters. Finally, for all $\jx\in[\sln]$ and $\dep\in\nat$ all vertices/arcs at depth $\dep$ in $\tree{\ix}{\jx}$ share their parameters (i.e. we have the same parameters across all $\jx\in[\sln]$ and vertices/arcs at depth $\dep$ in $\tree{\ix}{\jx}$). Note that this parameter sharing scheme means that, as in a transformer, we rely solely on the positional encodings to exploit the sequential structure (if our instances are sequences). For sequential instances we could exploit the sequential structure more by sharing parameters in $\tree{\ix}{\jx}$ as we did in Section \ref{tsenexsec}, although the positional encodings will still be required.

Finally, we construct a binary tree (as a subtree of $(\ver,\arc)$) whose leaves are the vertices $\lav{\nml}{\jx}$ for all $\jx\in[\sln]$. All vertices/arcs at the same depth in this tree share their parameters. The root of this tree is the root $\rot$ of the graph $(\ver,\arc)$.

The attention mechanism in the above exnet is \emp{single-headed}. However, we can generalise to \emp{multi-headed} attention. Here we have some $\nmh\in\nat$. The change to the above exnet is that, for all layers $\ix\in[\nml-1]$ and all $\jx\in[\sln]$, instead of having a single tree $\tree{\ix}{\jx}$ we instead have a sequence of trees $\seq{\tret{\ix}{\jx}{\hed}}{\hed\in[\nmh]}$. Each of these trees has, as its leaves, the set $\{\dob{\ix}{\jx}{\kx}\,|\,\kx\in[\sln]\}$. We note that each of these trees has different parameters (although for each $\hed\in[\nmh]$ the parameters are shared as above). Finally, for all layers $\ix\in[\nml-1]$ and all $\jx\in[\sln]$ we create an additional balanced binary tree $\htr{\ix}{\jx}$ whose leaves are the roots of the trees in $\seq{\tret{\ix}{\jx}{\hed}}{\hed\in[\nmh]}$ and whose root is the vertex $\lav{\ix+1}{\jx}$. We note that within the tree $\htr{\ix}{\jx}$ the vertices/arcs don't share parameters, but the parameters of the neural networks in the trees $\htr{\ix}{\jx}$ are the same for all $\jx$.

\subsection{Supernodes}

In the above exnets (the tree-structured exnets and the attention-based exnets), it could be the case that having a single primary extraction per vertex is not enough. We can hence replace every vertex in the above exnets by a \emp{supernode} which we now define. Specifically, given an exnet $(\ver,\arc)$, we can convert it into an exnet $(\ver',\arc')$ as follows. We have a \emp{width} $\wid\in\nat$. A supernode is a set of $\wid$ vertices in $\ver'$. For each vertex $v\in\ver$ we create a corresponding supernode in $\ver'$. Note that all these supernodes are disjoint. Now we add the following vertices and arcs to construct $(\ver',\arc')$. First, given any supernode $\spn$ corresponding to an internal vertex $v\in\ver\setminus\ips$, we let $\lch{\spn}$ and $\rch{\spn}$ denote the supernodes corresponding to $\lch{v}$ and $\rch{v}$ respectively. For each such supernode $\spn$ we create (as subgraphs of $(\ver,\arc)$), $\wid$ balanced binary tress. The set of leaves of each of these trees are the vertices in $\lch{\spn}\cup\rch{\spn}$. The roots of these trees are the elements of $\spn$. It is crucial that these trees, and the vertices within a tree, do not share parameters (although two supernodes can share parameters). Finally, we create a new balanced binary tree whose root will be the root of $(\ver',\arc')$ and whose leaves are the elements in the supernode corresponding to the root of $(\ver,\arc)$.

\section{An Alternative Algorithm}\label{altasec}

One potential issue with \alg\ is that, given some $v\in\ver\setminus\ips$ with $\lch{v}\notin\ips$, the parameters of the primary propagator on $\lch{v}$ may change too fast for the parameters of the primary propagator on $v$ to keep up. The same may be true for $\rch{v}$ as well as the complementary propagators. We hope that this is not actually an issue, but if it is we propose the following alternative algorithm called \alga. For simplicity here, we will assume that no parameters are shared. For \alga\ we assume that we have a finite training set $\trns\subseteq\ins\times\lss$, so that on each trial $t$ we have that $(\nint{t},\lost{t})$ is drawn uniformly at random from $\trns$.  Let $\trni$ be the set of all $\ain\in\ins$ such that there exists $\arl\in\lss$ with $(\ain,\arl)\in\trns$. We will maintain functions:
\be
\apf:\ver\times\trni\rightarrow\pss~~~~~;~~~~~\acf:\ver\times\trni\rightarrow\css
\ee
These functions are dynamic in that on some trials they will be updated. Also, parameters will only be modified when we say so. In the following description the functions are only updated when we say they are. For all $v\in\ips$ and $\ain\in\trni$ we will enforce that:
\be
\apf(v,\ain):=\tokf(\ain)(v)
\ee
We will also enforce that, for all $\ain\in\trni$, we have:
\be
\acf(\rot,\ain):=0
\ee
On each trial $t$ and for all $v\in\ver\setminus\ips$, the extractions $\apf(v,\nint{t})$ and $\acf(v,\nint{t})$ correspond to $\pme{t}{v}$ and $\cme{t}{v}$ in \alg\ respectively. \alga\ runs over epochs, where each epoch is a contiguous segment of trials. On each epoch we choose a vertex $v\in\ver\setminus\ips$ and do as follows. Given $t$ is the first trial of the epoch we, for all $z\in\pars{v}$, set the parameters $\ppi{t}{v}$, $\tpi{t}{v}$ and $\cpi{t}{z}{v}$ to be randomly chosen (initial) values. i.e. these neural networks are reset. Throughout the epoch we then update these parameters via stochastic gradient descent in order to minimise the expectation of:
\be
\lost{t}\left(\trr\left(\tpi{t}{v}, \ppr(\ppi{t}{v}, \apf(\lch{v},\nint{t}), \apf(\rch{v},\nint{t})), \sum_{z\in\pars{v}}\cpr(\cpi{t}{z}{v}, \acf(z,\nint{t}), \apf(\asi{z}{v},\nint{t}))\right)\right)
\ee
as in the deterministic mode of \alg. Given $t$ is the last trial of the epoch, we then (at the end of trial $t$) update the functions $\apf(v,\blank)$ and $\acf(v,\blank)$ such that for all $\ain\in\trni$ we have:
\be
\apf(v,\ain):=\ppr(\ppi{t}{v}, \apf(\lch{v},\ain), \apf(\rch{v},\ain))
\ee
\be
\acf(v,\ain):=\sum_{z\in\pars{v}}\cpr(\cpi{t}{z}{v}, \acf(z,\ain), \apf(\asi{z}{v},\ain))
\ee
An \emp{aeon} is composed of $|\ver\setminus\ips|$ epochs: one for each vertex $v\in\ver\setminus\ips$. The order of the vertex choices in an aeon is important: we run an epoch on a vertex $v$ only when an epoch has been run on $\lch{v}$ (or $\lch{v}\in\ips$) and an epoch has been run on $\rch{v}$ (or $\rch{v}\in\ips$). This ensures that the primary architecture at the end of an aeon is consistent. At the end of an aeon, and only at the end of an aeon, the primary architecture is consistent so can be used for prediction for new instances. \alga\ loops over many aeons. This completes the description of \alga. It is possible to incorporate shared parameters by updating all vertices that share specific parameters in the same epoch, as long as the required vertex order in each aeon can still be achieved. Note that a downside of \alga\ is the potentially large space complexity in order to store the functions $\apf$ and $\acf$.

\section*{Acknowledgements}

Research funded by the Defence Science and Technology Laboratory (Dstl) which is an executive agency of the UK Ministry of Defence providing world class expertise and delivering cutting-edge science and technology for the benefit of the nation and allies. The research supports the Autonomous Resilient Cyber Defence (ARCD) project within the Dstl Cyber Defence Enhancement programme.

\bibliographystyle{plain}
\bibliography{FNbib}

\end{document}